\documentclass[letterpaper, 10 pt, conference]{ieeeconf}  

\IEEEoverridecommandlockouts                              

\usepackage{graphics} 
\usepackage{amsmath} 
\usepackage{tikz}
\usepackage{import}
\usepackage{pgfplots}
\usepackage{graphicx}
\graphicspath{{./Imgs/}}
\usepackage{todonotes}
\usepackage[utf8]{inputenc}
\usepackage{array}
\usepackage{url}
\usepackage{gensymb}
\usepackage{upgreek}
\usepackage{booktabs}

\title{\LARGE \bf
High-speed Autonomous Racing using Trajectory-aided \\Deep Reinforcement Learning
}

\author{Benjamin Evans$^{1}$, Herman A. Engelbrecht$^{1}$ and Hendrik W. Jordaan$^{1}$
\thanks{$^{1}$Electrical and Electronic Engineering Department,
        Stellenbosch University, Stellenbosch, 7600, South Africa. 
        {\tt\small bdevans@sun.ac.za};
        {\tt\small hebrect@sun.ac.za};
        {\tt\small wjordaan@sun.ac.za}%
        }
}

\newcommand{\Fone}{F1Tenth }

\begin{document}

\maketitle
\thispagestyle{empty}
\pagestyle{empty}

\begin{abstract}

The classical method of autonomous racing uses real-time localisation to follow a precalculated optimal trajectory.
In contrast, end-to-end deep reinforcement learning (DRL) can train agents to race using only raw LiDAR scans.
While classical methods prioritise optimization for high-performance racing, DRL approaches have focused on low-performance contexts with little consideration of the speed profile.
This work addresses the problem of using end-to-end DRL agents for high-speed autonomous racing.
We present trajectory-aided learning (TAL) that trains DRL agents for high-performance racing by incorporating the optimal trajectory (racing line) into the learning formulation.
Our method is evaluated using the TD3 algorithm on four maps in the open-source F1Tenth simulator.
The results demonstrate that our method achieves a significantly higher lap completion rate at high speeds compared to the baseline.
This is due to TAL training the agent to select a feasible speed profile of slowing down in the corners and roughly tracking the optimal trajectory.

\end{abstract}

\section{Introduction}

Autonomous racing is a useful testbed for high-performance autonomous algorithms due to the nature of competition and the easy-to-measure performance metric of lap time \cite{Betz2022AutonomousRacing}.
The aim of autonomous racing is to use onboard sensors to calculate control references that move the vehicle around the track as quickly as possible.
Good racing performance operates the vehicle on the edge of its physical limits between going too slowly, which is poor racing behaviour, and going too fast, which results in the vehicle crashing.

The classical robotics approach uses control systems that depend on explicit state estimation to calculate references for the robot's actuators \cite{wang2021data}.
Classical racing systems use a localisation algorithm to determine the vehicle's pose on a map, which a path follower uses to track an optimal trajectory \cite{Heilmeier2020MinimumCar}. 
Methods requiring explicit state representation (localisation) are limited by requiring a map of the track and being inflexible to environmental changes  \cite{WalshCDDTLocalization}.

In contrast to classical methods, deep learning agents use a neural network to map raw sensor data (LiDAR scans) directly to control commands without requiring explicit state estimation \cite{zhang2016learning}.
Deep reinforcement learning (DRL) trains neural networks from experience to select actions that maximise a reward signal \cite{sutton2018reinforcement}. 
Previous DRL approaches have presented end-to-end solutions for \Fone racing but have been limited to low speeds \cite{hamilton2022zero, ivanov2020case}, and have lacked consideration of the speed profile \cite{Bosello2022TrainRaces}.

\begin{figure}[t]
    \centering
    \includegraphics[width=0.4\textwidth]{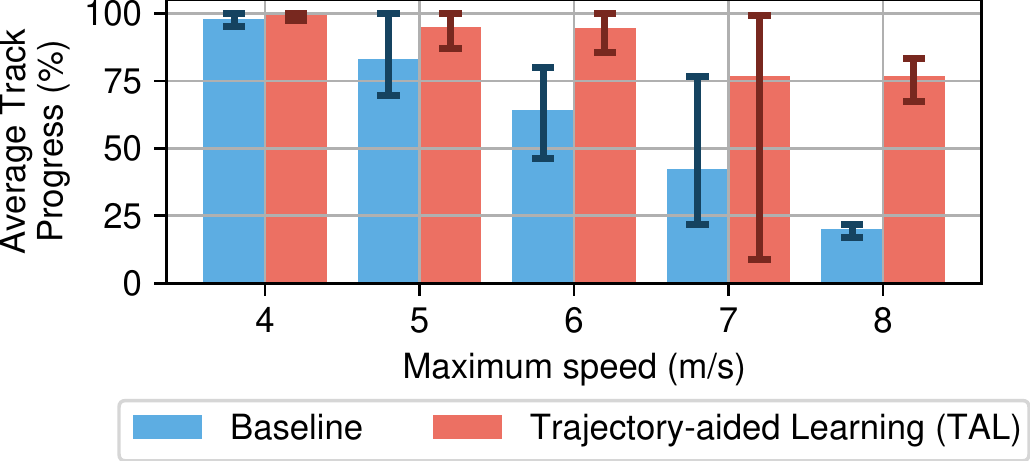}
    \caption{Our method achieves significantly higher average progress around the track at high speeds than the baseline.}
    \label{fig:progress_max_speed}
\end{figure}

This paper approaches the problem of how to train DRL agents for high-speed racing using only a LiDAR scan as input.
We provide insights on learning formulations for training DRL agents for high-performance control through the following contributions:
\begin{enumerate}
    \item Present trajectory-aided learning (TAL), which uses an optimal trajectory to train DRL agents for high-speed racing using raw LiDAR scans as input. 
    \item Demonstrate that TAL improves the completion rate of DRL agents at high speeds compared to the baseline learning formulation, as shown in Fig. \ref{fig:progress_max_speed}.
    \item Demonstrate that TAL agents select speed profiles similar to the optimal trajectory and outperform related approaches in the literature.
\end{enumerate}


\section{Literature Study}

\begin{figure}[h]
    \centering
    \includegraphics[width=0.44\textwidth]{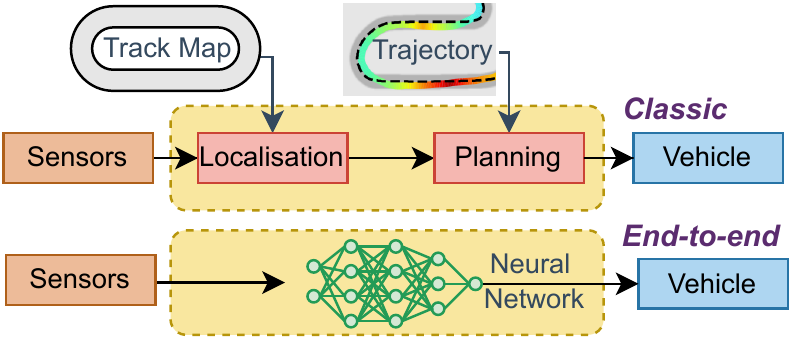}
    \caption{Classical racing stack using localisation and planning modules, and end-to-end racing using a neural network without state estimation.}
    \label{fig:classical_vs_end_to_end}
\end{figure}

We study methods of autonomous racing in the categories of classical methods and end-to-end learning.
Fig. \ref{fig:classical_vs_end_to_end} shows how the classical racing pipeline uses a localisation module to enable a planner to track a precomputed optimal trajectory, and end-to-end learning replaces the entire pipeline with a neural network-based agent.

\subsection{Classical Racing} \label{subsec:classic_racing}

The classical racing method calculates an optimal trajectory and then uses a path-following algorithm to track it \cite{Betz2022AutonomousRacing}.
Trajectory optimisation techniques calculate a set of waypoints (positions with a speed reference) on a track that, when followed, lead the vehicle to complete a lap in the shortest time possible \cite{Heilmeier2020MinimumCar}.
A path-following algorithm tracks the trajectory using the vehicle's pose as calculated by a localisation algorithm.

\textbf{Localisation:} 
Localisation approaches for autonomous racing depend on the sensors and computation available.
Full-sized racing cars are often equipped with GPS (GNSS), LiDAR, radar, cameras, IMUs, and powerful computers that can fuse these measurements in real-time \cite{wischnewski2022indy}.
Classical \Fone racing approaches have used a particle filter that takes a LiDAR scan and a map of the track to estimate the vehicle's pose \cite{WalshCDDTLocalization, o2020tunercar, wang2021data}.
Localisation methods are inherently limited by requiring a race track map and, thus, are inflexible to unmapped tracks.

\textbf{Classical Path-Following:} 
Model-predictive controllers (MPC) and pure pursuit path-followers have been used for trajectory tracking \cite{Betz2022AutonomousRacing}.
MPC planners calculate optimal control commands in a receding horizon manner \cite{tuatulea2020design} and have demonstrated high-performance results racing \Fone vehicles at speeds of up to 7 m/s \cite{wang2021data}.
The pure pursuit algorithm uses a geometric model to calculate a steering angle to follow the optimal trajectory \cite{coulter1992implementation}, and has been used to race at speeds of 7 m/s \cite{o2020tunercar} and over 8 m/s \cite{becker2022model}.

\textbf{Learning-based Path-following:}
Classical path-following algorithms have been replaced by neural networks, aiming to improve computational efficiency (compared to MPC) \cite{tuatulea2020design, chisari2021learning} and performance in difficult-to-model conditions such as drifting \cite{Cai2020HighSpeedLearning}.
Including upcoming trajectory points in the state vector (as opposed to only centerline points \cite{chisari2021learning}) has shown to improve racing performance \cite{ghignone2022tc, dwivedi2022continuous}.
This shows demonstrates that using the optimal trajectory results in high-performance racing.

While classical and learning-based path-following methods have produced high-performance results, they are inherently limited by requiring the vehicle's location on the map.

\subsection{End-to-end Learning} \label{subsec:e2e_learning}

In contrast to classical methods that use a perception, planning and control pipeline, end-to-end methods use a neural network to map raw sensory data to control references \cite{Bosello2022TrainRaces}.
While some approaches have used camera images \cite{jaritz2018end}, the dominant input has been LiDAR scans \cite{hamilton2022zero, Bosello2022TrainRaces, evans2022accelerating}.

\textbf{Autonomous Driving:} 
End-to-end learning agents can use a subset of beams from a LiDAR scan to output steering references that control a vehicle travelling at constant speed \cite{hamilton2022zero}.
While imitation learning (IL) has been used to train agents to copy an expert policy \cite{sun2022benchmark}, deep reinforcement learning, has shown better results, with higher lap completion rates \cite{hamilton2022zero}.
DRL algorithms train agents in an environment (simulation \cite{hamilton2022zero} or real-world system \cite{evans2022accelerating}), where at each timestep, the agent receives a state, selects an action and then receives a reward.
DRL approaches to driving \Fone vehicles have considered low, constant speeds of 1.5 m/s \cite{hamilton2022zero, musau2022using}, 2 m/s \cite{evans2022accelerating}, and 2.4 m/s \cite{ivanov2020case}.
While indicating that DRL agents can control a vehicle, these methods neglect the central racing challenge of speed selection.

\textbf{Autonomous Racing:} 
Using model-free end-to-end DRL agents to select speed and steering commands for autonomous racing is a difficult problem \cite{Brunnbauer2022LatentRacing, zhang2022residual}.
In response, Brunnbauer et al. \cite{Brunnbauer2022LatentRacing} turned to model-based learning and Zhang et al. \cite{zhang2022residual} incorporated an artificial potential field planner in the learning to simplify the learning problem.
Both \cite{Brunnbauer2022LatentRacing} and \cite{zhang2022residual} show that their agents regularly crash while using top speeds of only 5 m/s, demonstrating the difficulty of learning for high-speed autonomous racing.
Bosello et al. \cite{Bosello2022TrainRaces} use a model-free DRL algorithm (DQN) for \Fone racing at speeds of up to 5 m/s, but provide no detail on the speed profile, trajectory or crash rate.

\textbf{Summary:}
Classical racing methods have produced high-performance racing behaviour using high maximum speeds but are limited by requiring localisation.
In contrast, end-to-end DRL agents are successful in controlling vehicles at low speeds using only the LiDAR scan as input.
While some methods have approached speed selection using DRL agents, there has been little study on the speed profiles selected, and the highest speed used is 5~m/s, which is significantly less than classical methods of 8~m/s.
This paper targets the gap in developing high-performance racing solutions for steering and speed control in autonomous race cars.

\section{Methodology}

\subsection{Reinforcement Learning Preliminary}

Deep reinforcement learning (DRL) trains autonomous agents, consisting of deep neural networks, to maximise a reward signal from experience \cite{sutton2018reinforcement}.
Reinforcement learning problems are modelled as Markov Decision Processes (MDPs), where the agent receives a state $s$ from the environment and selects an action $a$.
After each action has been executed, the environment returns a reward $r$ indicating how good or bad the action was and a new state $s'$.

This work considers deep-deterministic-policy-gradient (DDPG) algorithms since we work with a continuous action space \cite{Lillicrap2016ContinuousLearning}.
DDPG algorithms maintain two neural networks, an actor $\mu$ that maps a state to an action and a critic $Q$ that evaluates the action-value function.
A pair of networks are maintained for the actor and the critic; the model networks are used to select actions, and target networks calculate the targets $\mu'$ and $Q'$.
A replay memory collects the agent's experience of acting and receiving rewards. 
After each step, a batch of $N$ transitions is randomly sampled from memory and used to update the networks.

The critic is trained to learn the Q-value for each state-action pair $Q(s, a)$.
For each transition, $j$ in the batch, the bootstrapped target  $y_j$ is calculated using the Bellman equation by adding the reward earned and the discounted Q-value for the next state if the agent follows its target policy.
The actor, parameterised by $\theta$, is trained to maximise the objective $J(\theta)$ of selecting actions with high Q-values.
The gradient that maximises the objective $J(\theta)$ is calculated as,
\begin{equation}
     \nabla_{\theta} J(\theta ) = \frac{1}{N} \sum_j  \nabla_{\theta} Q(s_j, \mu(s_j)).
    \label{eqn:ddpg_policy_grad}
\end{equation}
After each network update, a soft update is applied to adjust the target networks towards the model networks.

The twin-delayed-DDPG (TD3) algorithm improves the original DDPG algorithm by using a pair of Q-networks and smoothing the policy by adding noise to the actions selected by target policy~\cite{fujimoto2018addressing}.
The TD3 Q-targets are calculated using the minimum of the pair of Q-networks,
\begin{equation}
    \begin{split}
        y_j  = & r_j + \gamma \min_{i=1,2} Q_i' (s_j', \mu' (s_j') + \epsilon)\\
         & \epsilon \sim \text{clip} (\mathcal{N} (0, \sigma), -c, c).\\
    \end{split}
\end{equation}
In the equation, $\gamma$ is the discount factor, $i$ is the number of the Q-network (i.e. $Q_1', Q_2'$), $\mu'$ is the target actor network, $\epsilon$ is the clipped noise sampled from the normal distribution $\mathcal{N}$, and $c$ is the noise clipping constant.
The TD3 algorithm introduces delayed policy updates by only updating the policy network after every second Q-network update.

\subsection{End-to-end Learning Problem Formulation}

\begin{figure}[h]
    \centering
    \includegraphics[width=0.48\textwidth]{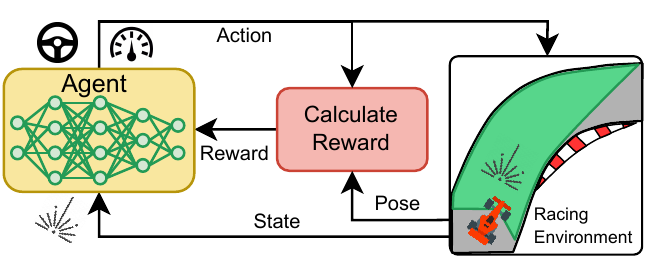}
    \caption{The DRL agent receives a state, selects an action that is implemented, and a reward based on the agent's action and vehicle's position is calculated and given to the agent.}
    \label{fig:drl_problem}
\end{figure}

End-to-end learning replaces the entire processing pipeline with a learning agent. 
The input to the agent is a state vector representing the environment, and the output is an action vector used to control the vehicle.
Fig. \ref{fig:drl_problem} shows the flow of information with the agent receiving a state consisting of the LiDAR scan and selecting an action of a speed and steering angle.
A reward is calculated based on the agent's action and the vehicle's pose in the environment.

\textbf{State Vector:}
The agent uses a state vector of 20 evenly spaced beams from the LiDAR scan with a field of view of $\uppi$ radians.
The LiDAR scans from the previous and current planning steps are stacked together so that the agent can infer the vehicle's speed.
Each beam is scaled according to the maximum of 10 m, resulting in values between 0 and 1 used as input into the neural network.

\textbf{Action Vector:}
The agent outputs two continuous actions in the range $[-1, 1]$, which are used for the two control variables of steering angle and speed.
The steering action is scaled according to the maximum steering angle, and
the speed is scaled to the range $[1, v_\text{max}]$ m/s, where $v_\text{max}$ is the maximum speed.
The minimum speed of 1 m/s is prevents the vehicle from not moving.

\subsection{Trajectory-aided Learning}

We present trajectory-aided learning (TAL), a reward signal that trains an agent to follow the optimal trajectory.
Our approach is motivated by the literature showing that classical solutions using trajectory optimisation and path-following approaches achieve high-performance racing \cite{wischnewski2022indy, becker2022model}.
While imitation learning from expert data (including from a pure pursuit expert \cite{sun2022benchmark}) has demonstrated poor lap completion results \cite{hamilton2022zero}, deep reinforcement learning has successfully trained agents to race \cite{Bosello2022TrainRaces}.
Therefore, we propose incorporating a classical solution in the DRL reward signal to train end-to-end agents for high-performance racing.

\begin{figure}[h]
    \centering
    \includegraphics[width=0.85\linewidth]{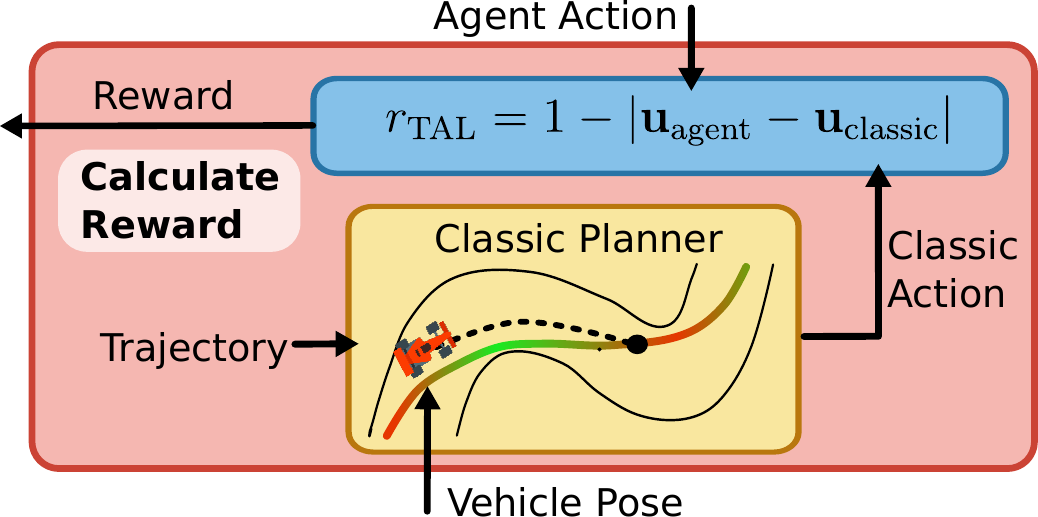}
    \caption{The trajectory-aided learning reward is calculated using the difference between the agent action $\mathbf{u}_\text{agent}$ and classic planner action $\mathbf{u}_\text{classic}$.}
    \label{fig:tal_architecture}
\end{figure}

\textbf{TAL Reward:}
The reward signal should train the agent to drive as fast as possible while maintaining safety and not crashing.
A base reward of giving a punishment of -1 for crashing and a reward of 1 for lap completion is combined with a shaped reward that encourages high-performance racing.
Fig. \ref{fig:tal_architecture} shows how the shaped trajectory-aided learning reward is calculated using the difference between the agent action and the action that a classic planner would have selected.
We write the reward as, 
\begin{equation}
        r_\text{TAL} = 1 - | v_\text{agent} - v_\text{classic}  | - | \delta_\text{agent} - \delta_\text{classic} |,
\end{equation}
where $v$ represents the speed and $\delta$ the steering angle.
In this equation, the subscript ``classic'' refers to the actions the classical planner would select, and the subscript ``agent'' refers to the action selected by the agent.
The shaped reward is scaled by 0.2 and clipped to be above 0.


\textbf{Classical Planner:}
The high-performance behaviour of the classic planner is a guide for the learning agent.
Fig. \ref{fig:tal_architecture} shows how the classic planner action is calculated using the vehicle pose, optimal trajectory and a path-following algorithm.
The classical planner uses the trajectory optimisation method presented by Heilmeier et al. \cite{Heilmeier2020MinimumCar} to calculate a minimum curvature path with a minimum time speed profile.
The pure pursuit path-following algorithm \cite{coulter1992implementation} is used to track the optimal trajectory.
The classical planner selects the speed of the upcoming way-point as its speed action.

\subsection{Baseline Learning Formulation} \label{subsec:baseline_learning_formulation}

We compare our approach to a baseline reward encouraging the vehicle to track the centre line.
The baseline retains the standard reward of 1 for completing a lap and -1 for crashing.
At each step, a cross-track and heading reward is given to the agent to reward velocity in the track direction and punish lateral deviation \cite{jaritz2018end}.
The reward is written as,
\begin{equation}
    r_\text{baseline} = \frac{v_\text{t}}{v_\text{max}} \cos \psi -  ~ d_\text{c},
\end{equation}
where $v_\text{t}$ is the vehicle's speed, $v_\text{max}$ is the maximum speed, $\psi$ is the heading error angle, and $d_c$ is the cross-track distance.



\section{Evaluation}

\subsection{Experiment Design}

\begin{figure}[h]
    \centering
    \begin{tabular}{cccc}
      \includegraphics[width=1.2cm]{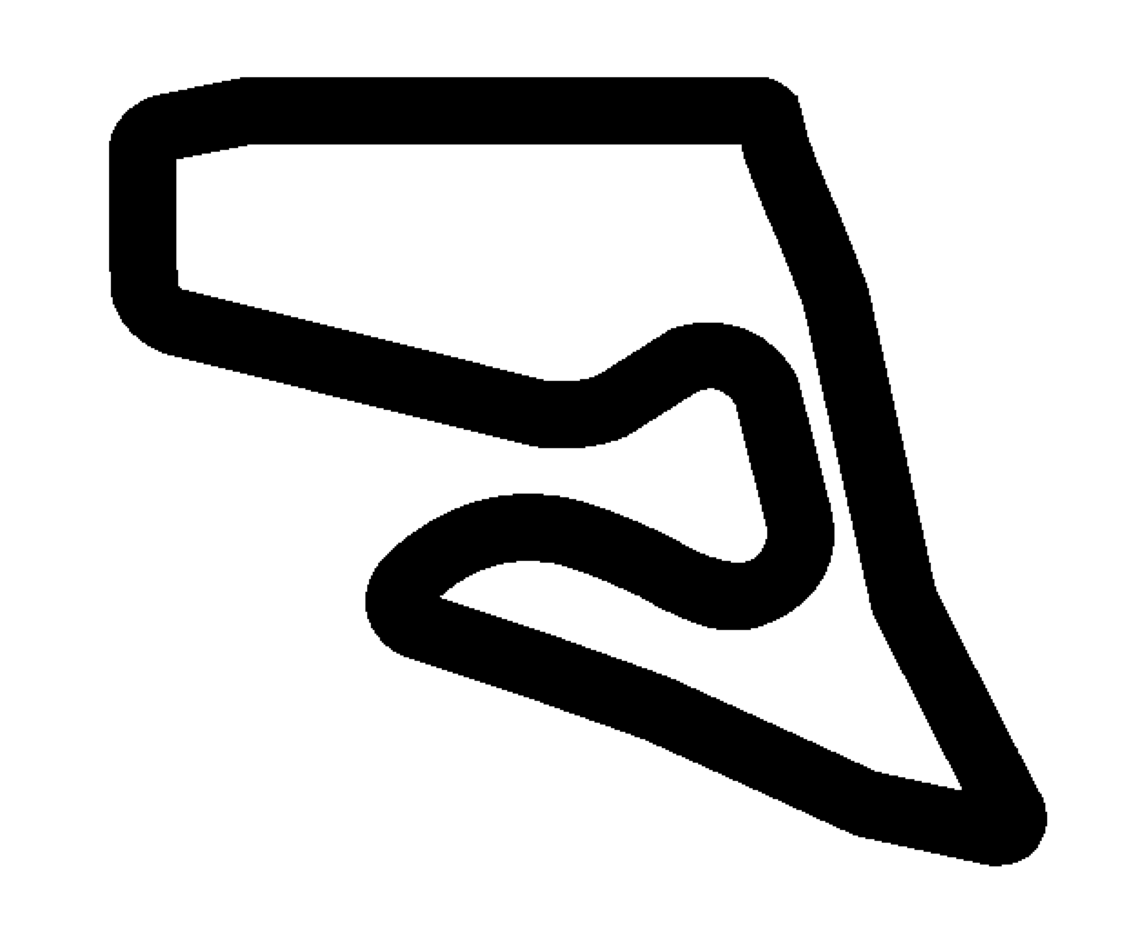}  & \includegraphics[width=1.9cm]{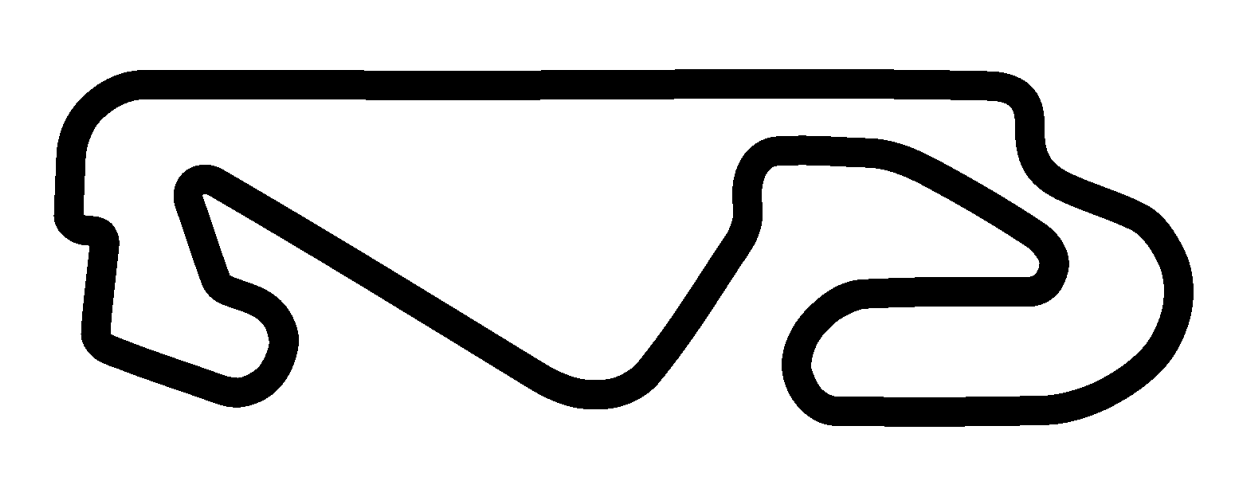}   & \includegraphics[width=1.8cm]{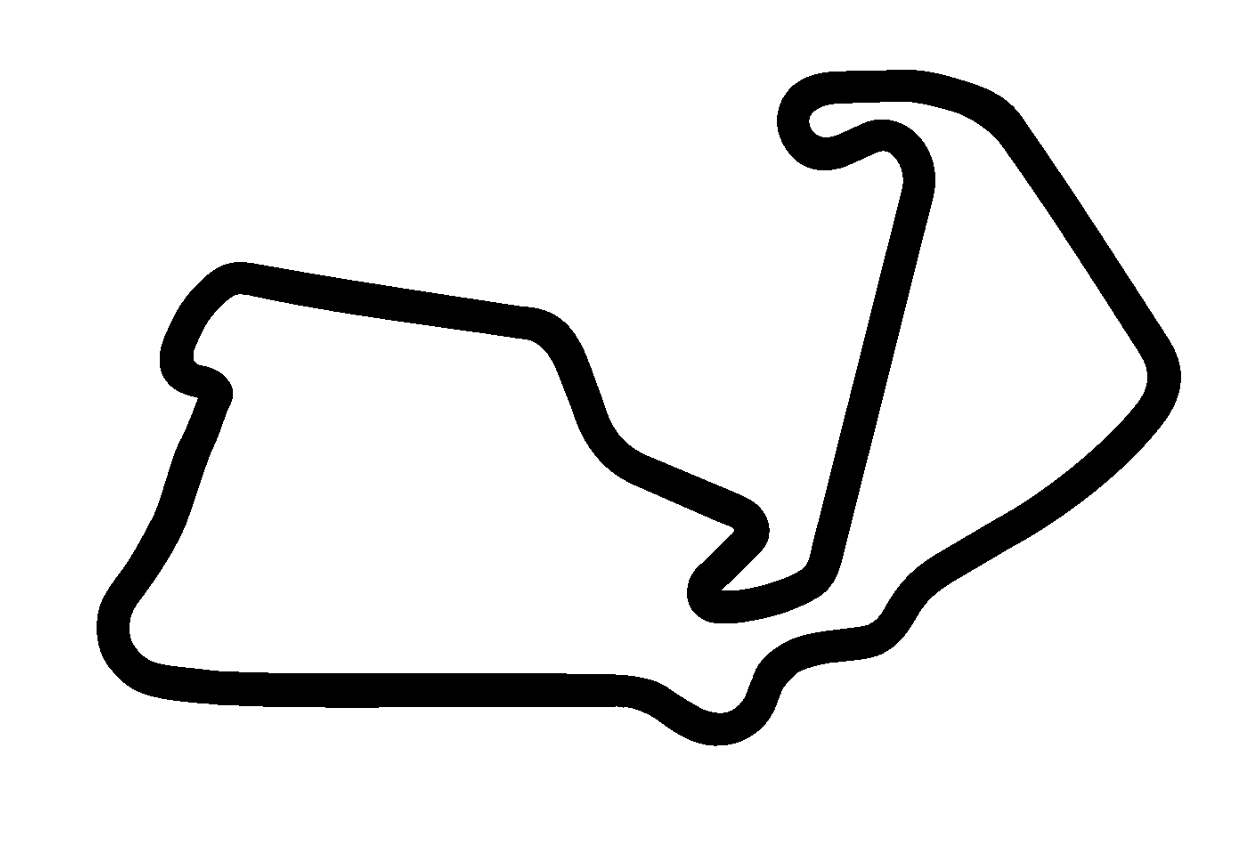}  & \includegraphics[width=1.7cm]{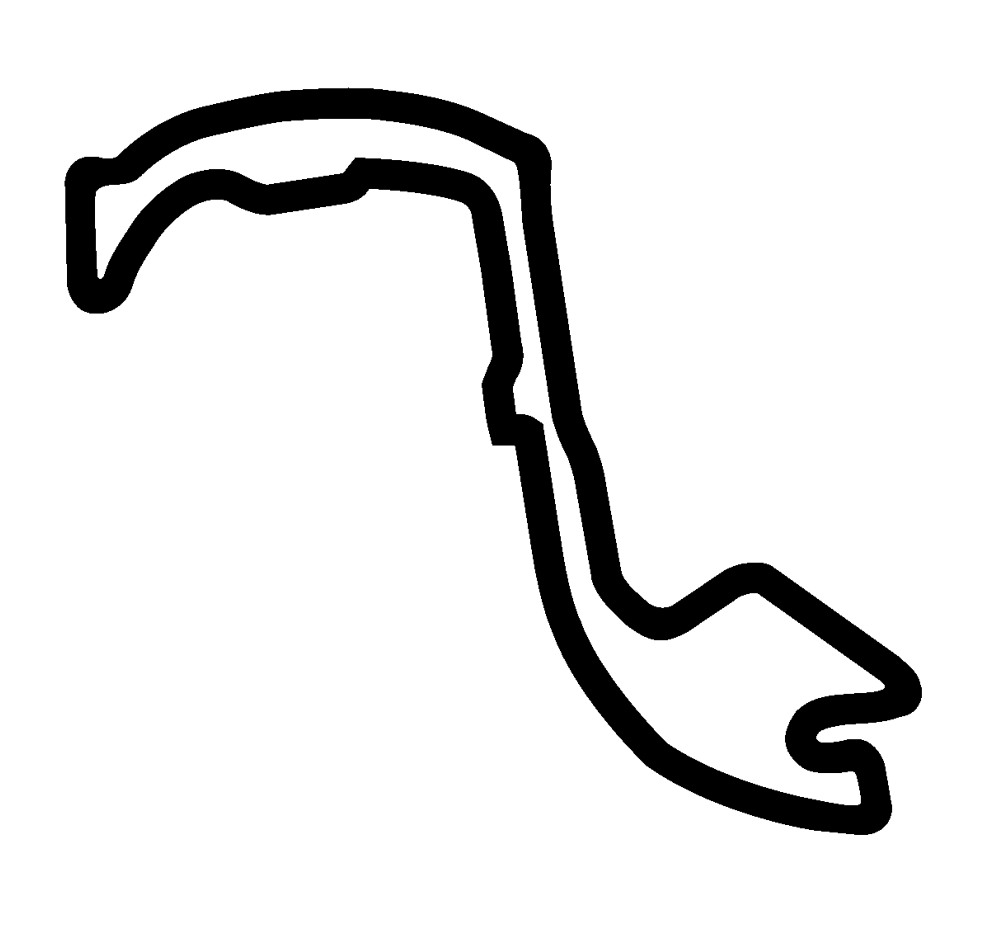}  \\
    \end{tabular}
    \caption{Map shapes of the AUT, ESP, GBR and MCO (left to right) tracks.}
    \label{fig:map_description}
\end{figure}

We evaluate our approach using the open-source \Fone simulator in \cite{o2020f1tenth}.
The simulator is modelled on the Gym style environments with a step method that takes an action and returns a state. 
The LiDAR scan is simulated using a ray-casting algorithm, and noise with a standard deviation of 0.01 is added to each beam.
Planning in the simulator takes place at 10 Hz, while the internal dynamics updates at 100 Hz. 
Fig. \ref{fig:map_description} shows the shapes of the four training maps, AUT, ESP, GBR and MCO, that are used in the evaluation.

\begin{figure}[h]
    \centering
    \def\svgwidth{0.8\linewidth}
\begingroup%
  \makeatletter%
  \providecommand\color[2][]{%
    \errmessage{(Inkscape) Color is used for the text in Inkscape, but the package 'color.sty' is not loaded}%
    \renewcommand\color[2][]{}%
  }%
  \providecommand\transparent[1]{%
    \errmessage{(Inkscape) Transparency is used (non-zero) for the text in Inkscape, but the package 'transparent.sty' is not loaded}%
    \renewcommand\transparent[1]{}%
  }%
  \providecommand\rotatebox[2]{#2}%
  \newcommand*\fsize{\dimexpr\f@size pt\relax}%
  \newcommand*\lineheight[1]{\fontsize{\fsize}{#1\fsize}\selectfont}%
  \ifx\svgwidth\undefined%
    \setlength{\unitlength}{411.1970437bp}%
    \ifx\svgscale\undefined%
      \relax%
    \else%
      \setlength{\unitlength}{\unitlength * \real{\svgscale}}%
    \fi%
  \else%
    \setlength{\unitlength}{\svgwidth}%
  \fi%
  \global\let\svgwidth\undefined%
  \global\let\svgscale\undefined%
  \makeatother%
  \begin{picture}(1,0.35237771)%
    \lineheight{1}%
    \setlength\tabcolsep{0pt}%
    \put(0,0){\includegraphics[width=\unitlength,page=1]{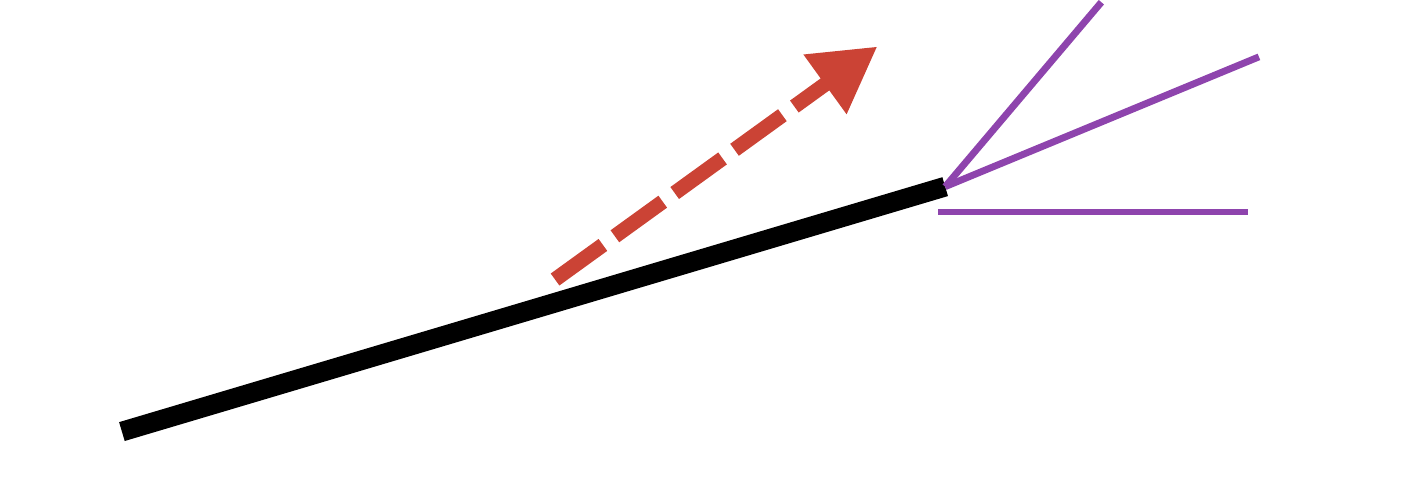}}%
    \put(0.83765019,0.23054793){\color[rgb]{0,0,0}\makebox(0,0)[lt]{\lineheight{1.25}\smash{\begin{tabular}[t]{l}$\theta$\end{tabular}}}}%
    \put(0.76039513,0.29239778){\color[rgb]{0,0,0}\makebox(0,0)[lt]{\lineheight{1.25}\smash{\begin{tabular}[t]{l}$\delta$\end{tabular}}}}%
    \put(0.53281223,0.21232211){\color[rgb]{0,0,0}\makebox(0,0)[lt]{\lineheight{1.25}\smash{\begin{tabular}[t]{l}$\beta$\end{tabular}}}}%
    \put(0.47845248,0.29959047){\color[rgb]{0,0,0}\makebox(0,0)[lt]{\lineheight{1.25}\smash{\begin{tabular}[t]{l}$v$\end{tabular}}}}%
    \put(0,0){\includegraphics[width=\unitlength,page=2]{bicycleModel_pdf.pdf}}%
    \put(0.21222585,0.15271738){\makebox(0,0)[lt]{\lineheight{1.25}\smash{\begin{tabular}[t]{l}$x$\end{tabular}}}}%
    \put(0.04773471,0.3108394){\makebox(0,0)[lt]{\lineheight{1.25}\smash{\begin{tabular}[t]{l}$y$\end{tabular}}}}%
    \put(0,0){\includegraphics[width=\unitlength,page=3]{bicycleModel_pdf.pdf}}%
  \end{picture}%
\endgroup%

    \caption{Single-track bicycle model used by the F1Tenth simulator.}
    \label{fig:single_track_model}
\end{figure}

\textbf{Vehicle Model:}
The simulator represents the vehicle using the kinematic bicycle model \cite{Althoff2017}.
Fig. \ref{fig:single_track_model} shows the model representing the vehicle with the state variables of position $x, y$, speed $v$, orientation (yaw) $\theta$, yaw rate $\dot \theta$ steering angle $\delta$ and slip angle $\beta$.
The 7-dimensional state is updated using the single-track bicycle model equations presented in \cite{Althoff2017}.
The model takes the parameters of vehicle mass, wheelbase length, height, cornering stiffness, coefficient of friction and moment of inertia.
The single-track model assumes a linear relationship between the slip angle and the lateral force, resulting in the model being accurate for small slip angles ($\approx < 8\degree$) but inaccurate for higher slip angles.

\textbf{Learning Implementation:}
The experiments use neural networks with two hidden layers of 100 neurons each. 
The \textit{ReLU} activation function is used after each hidden layer, and the \textit{tanh} function for the output layer to scale the output to the range [-1, 1].
The TD3 algorithm uses the Adam optimiser with a learning rate of 0.001, a batch size of 100, a discount factor of 0.99, exploration noise of 0.1, action smoothing noise of 0.2 and noise clipping at 0.5.

\textbf{Experiments:}
The evaluation compares the ability of the baseline (\S \ref{subsec:baseline_learning_formulation}) and TAL learning formulations to train DRL agents to race at high speeds through four experiments,
\begin{enumerate}
    \item Investigating the effect of maximum speeds ranging from 4 m/s to 8 m/s on performance.
    \item Comparing the lap times and completion rates of agents with a maximum speed of 6 m/s on training maps and tracks unseen during training.
    \item Comparing the trajectories, speed profiles and slip angles of agents with a maximum speed of 6 m/s.
    \item Comparing the TAL agent performance with a maximum speed of 8 m/s, to the classical method and competitive methods in the literature.
\end{enumerate}
The agents are trained for 100,000 steps in the simulator and tested by taking an average of 20 test laps.
All learning experiments are repeated five times with unique random seeds.
All the code from the experiments is seeded and available in the associated repository: \url{https://github.com/BDEvan5/TrajectoryAidedLearning}.

\subsection{Maximum Speed Investigation}

\begin{figure}[h]
    \centering
    \includegraphics[width=0.9\linewidth]{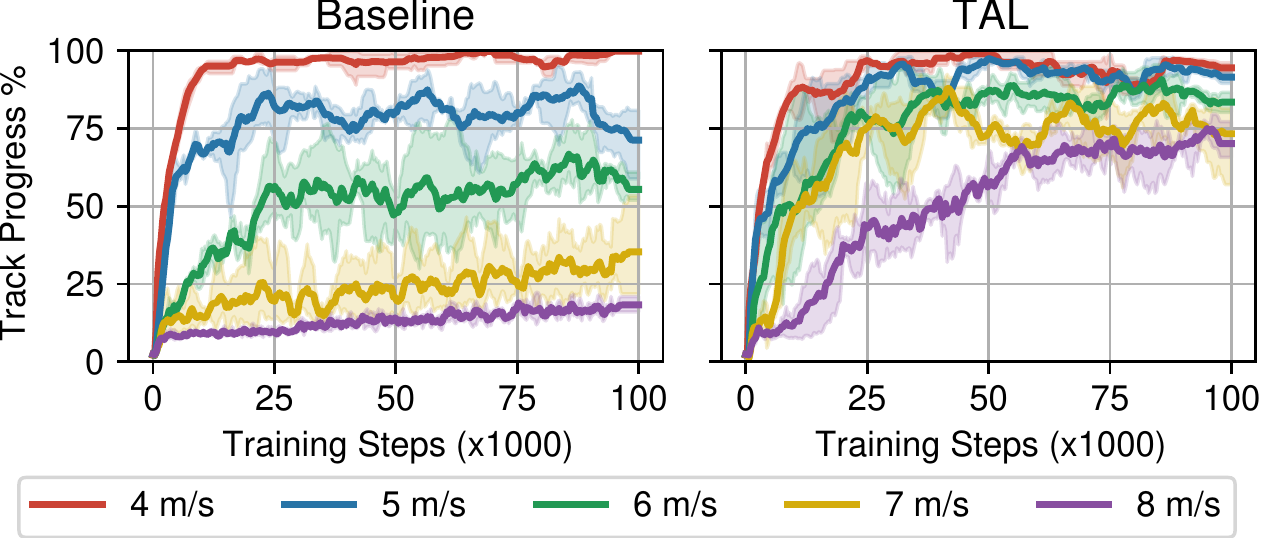}
    \caption{The average progress during training of the baseline and TAL agents on the ESP map. }
    \label{fig:training_progress}
\end{figure}

The first experiment investigates the effect of maximum speed on agent performance by training agents with increasing maximum speeds on the ESP map.
Fig. \ref{fig:training_progress} shows the average progress during training of the baseline and TAL agents.
The lines represent the average, and the shaded regions indicate the minimums and maximums of the middle three repeats.
The baseline graph shows that for a maximum speed of 4 m/s, the agent quickly learns to achieve average progress near 100\%.
As the maximum speed increases, the average progress decreases.
At 8 m/s, the average progress remains below 25\% for the entirety of the training.

In Fig. \ref{fig:training_progress}, the TAL agent's graph (right) shows that for all the maximum speeds considered, the agent learns to achieve over 75\% average progress.
The 6 m/s, 7 m/s and 8 m/s runs achieved averages of 80\%, 75\%, 70\% respectively.
The TAL agent's higher average progress shows an advantage over the baseline of travelling further without crashing.

\begin{figure}[h]
    \centering
    \vspace{2mm}
    \includegraphics[width=0.9\linewidth]{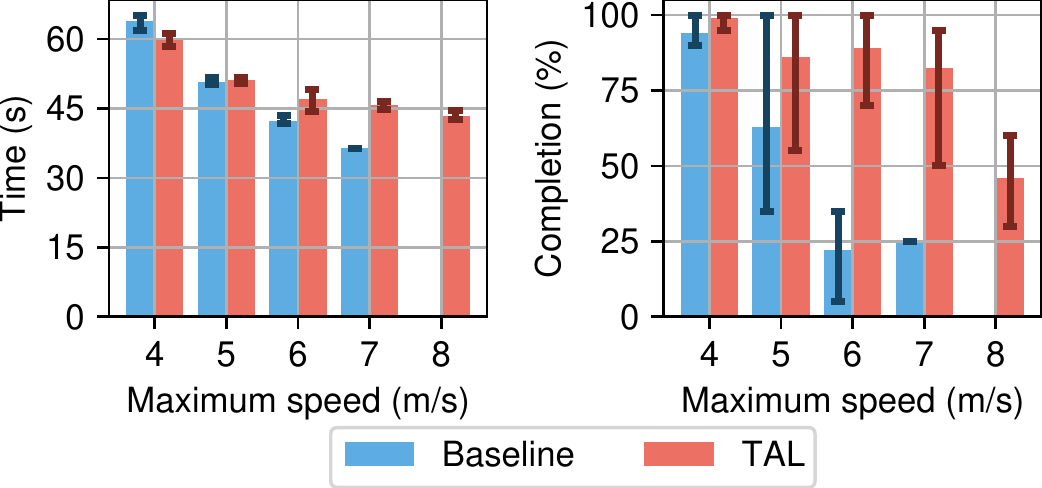}
    \caption{Lap times and completion rate of trained baseline and TAL agents on the ESP map.}
    \label{fig:trained_perfromance_tal_baseline}
\end{figure}

The lap times and completion rates of the trained baseline and TAL agents are plotted in Fig. \ref{fig:trained_perfromance_tal_baseline}.
The TAL agent has faster laps times for lower maximum speeds than the baseline.
As the maximum speed increases, the times even out and then the baseline achieves faster lap times than the TAL agent.

In Fig. \ref{fig:trained_perfromance_tal_baseline}, the completion graph (right) shows that the baseline agent completion rate starts at 100\% for the 4 m/s and drops off to 50\% for the 6 m/s and the 8 m/s agents do not complete any laps.
In contrast, the TAL agents all achieve higher completion rates, with the 6 m/s agent achieving a 60\% completion rate and the 8 m/s 40\%.
This is a similar result to the average progress shown in Fig. \ref{fig:progress_max_speed}
While the TAL agents also have lower completion rates at higher speeds, the results indicate a significant improvement over the baseline.

\subsection{Quantitative Performance Evaluation - 6 m/s}

The performance of the baseline and TAL agents is compared using a maximum speed of 6 m/s, since the baseline performs poorly at higher speeds.

\begin{figure}[h]
    \centering
    \includegraphics[width=0.9\linewidth]{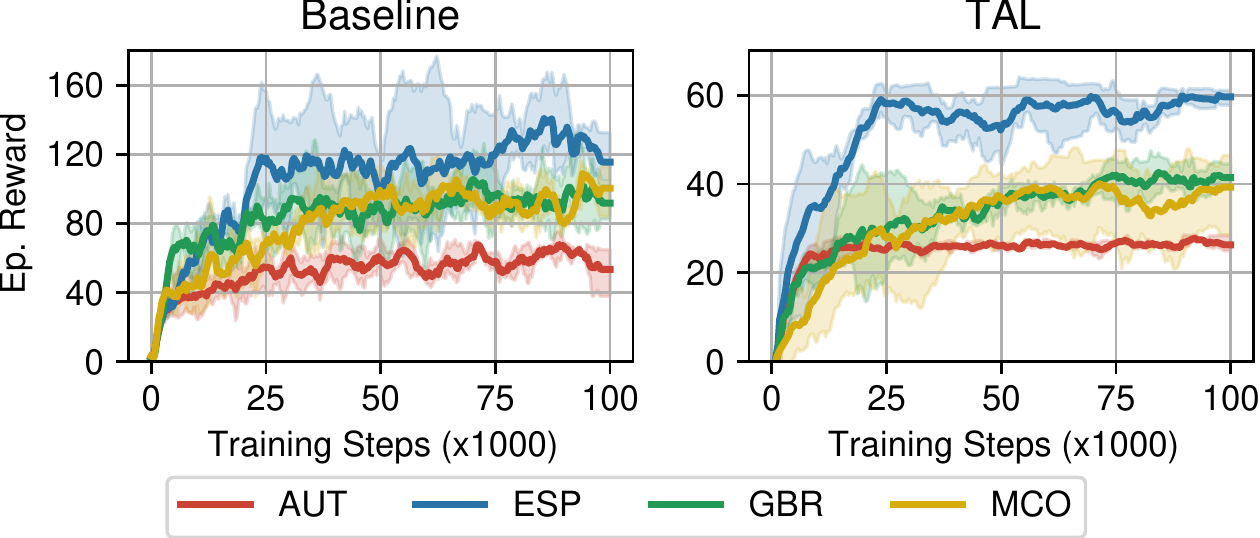}
    \caption{Episode rewards earned by training the baseline and TAL agents with a maximum speed of 6 m/s on the AUT, ESP, GBR and MCO maps.}
    \label{fig:Cth_TAL_maps_RewardTrainingGraph}
\end{figure}

Fig. \ref{fig:Cth_TAL_maps_RewardTrainingGraph} shows the episode rewards earned by the agents training them on the AUT, ESP, GBR and MCO maps.
The agents initially earn close to zero reward since the crash quickly.
The rewards across maps in both graphs show a similar trend of the agents achieving higher rewards on the longer ESP track (236.8 m), intermediate rewards for the GBR and MCO tracks (202.2 m and 178.3 m) and lower rewards for the shorter AUT track (93.7 m). 
The baseline reward signal provides larger rewards per episode than the TAL agent due to the scaling used in the calculation.

\begin{figure}[h]
    \centering
    \vspace{2mm}
    \includegraphics[width=0.9\linewidth]{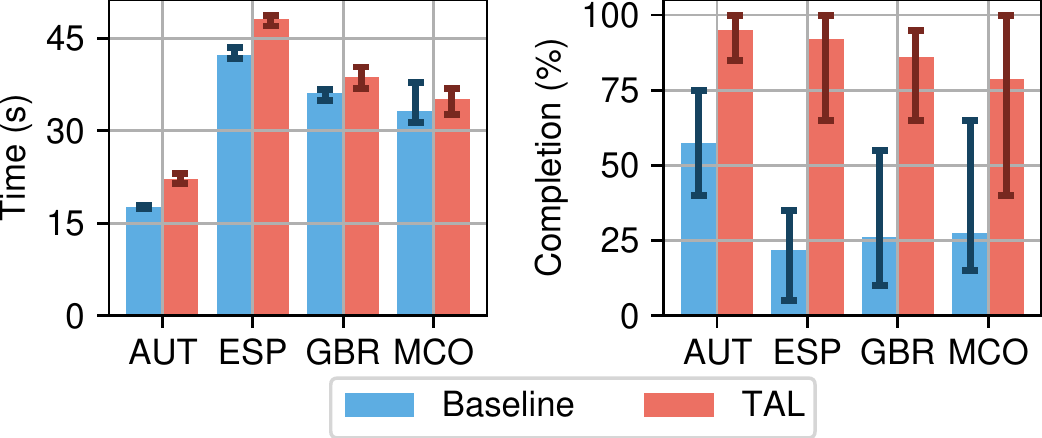}
    \caption{Average lap times and completion rates for the baseline, TAL and classical planners with a maximum speed of 6 m/s.}
    \label{fig:test_maps_baseline_tal}
\end{figure}

Fig. \ref{fig:test_maps_baseline_tal} shows the average lap times and completion rates for the classical, baseline and TAL planners with a maximum speed of 6 m/s.
While the baseline agent achieves slightly lower lap times than the TAL agent, the baseline agent has a significantly lower completion rate.
On the ESP, GBR and MCO maps, the baseline agent completes less than 25\% of the laps.
In contrast, the TAL agent completes over 75\% of the laps on all the tracks.
This result demonstrates that the TAL formulation results in agents achieving higher completion rates when using a maximum speed of 6 m/s.

\begin{figure}[h]
    \centering
    \includegraphics[width=0.9\linewidth]{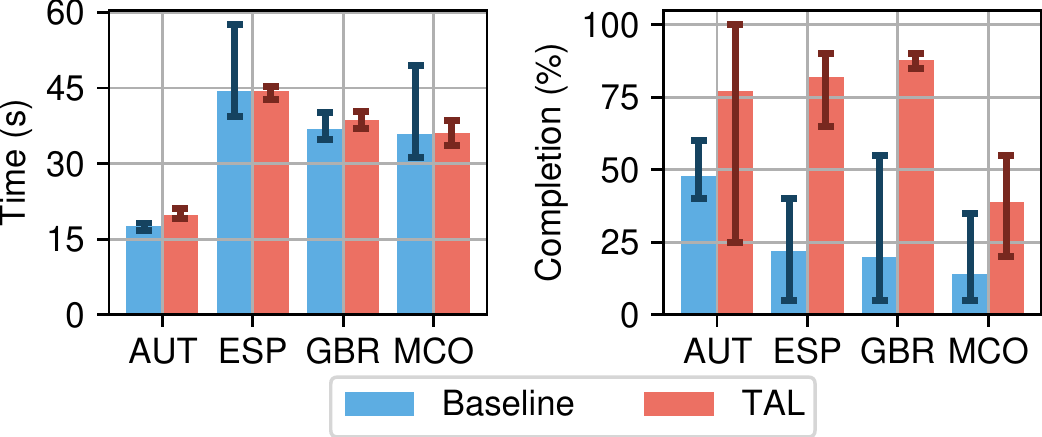}
    \caption{Lap times and completion rates for agents trained on the GBR map and tested on the AUT, ESP, GBR and MCO maps.}
    \label{fig:map_policy_generality}
\end{figure}

The generality of the learned policies is evaluated by testing the agents trained on the GBR track on all the test tracks.
Fig. \ref{fig:map_policy_generality} shows the lap times achieved by the baseline and TAL agents are close together, with the baseline agent having a larger deviation on the ESP and MCO tracks.

Fig. \ref{fig:map_policy_generality} shows that the TAL agent achieves significantly higher completion rates than the baseline on all the tracks.
The completion rates are all lower than when the agents were tested on the training track (Fig. \ref{fig:test_maps_baseline_tal}), indicating that while the policies learned do generalise to other tracks, there is a performance drop in the completion rate.
The TAL agent achieving significantly higher completion rates than the baseline agent, when tested on other maps, indicates that the TAL performance improvement is robust to different tracks.

\subsection{Qualitative Trajectory Analysis - 6 m/s}


We investigate the performance difference by comparing the trajectories of the baseline and TAL agents.


\begin{figure}[h]
    \centering
    \begin{minipage}{0.32\linewidth}
        \centering
        \includegraphics[width=\textwidth]{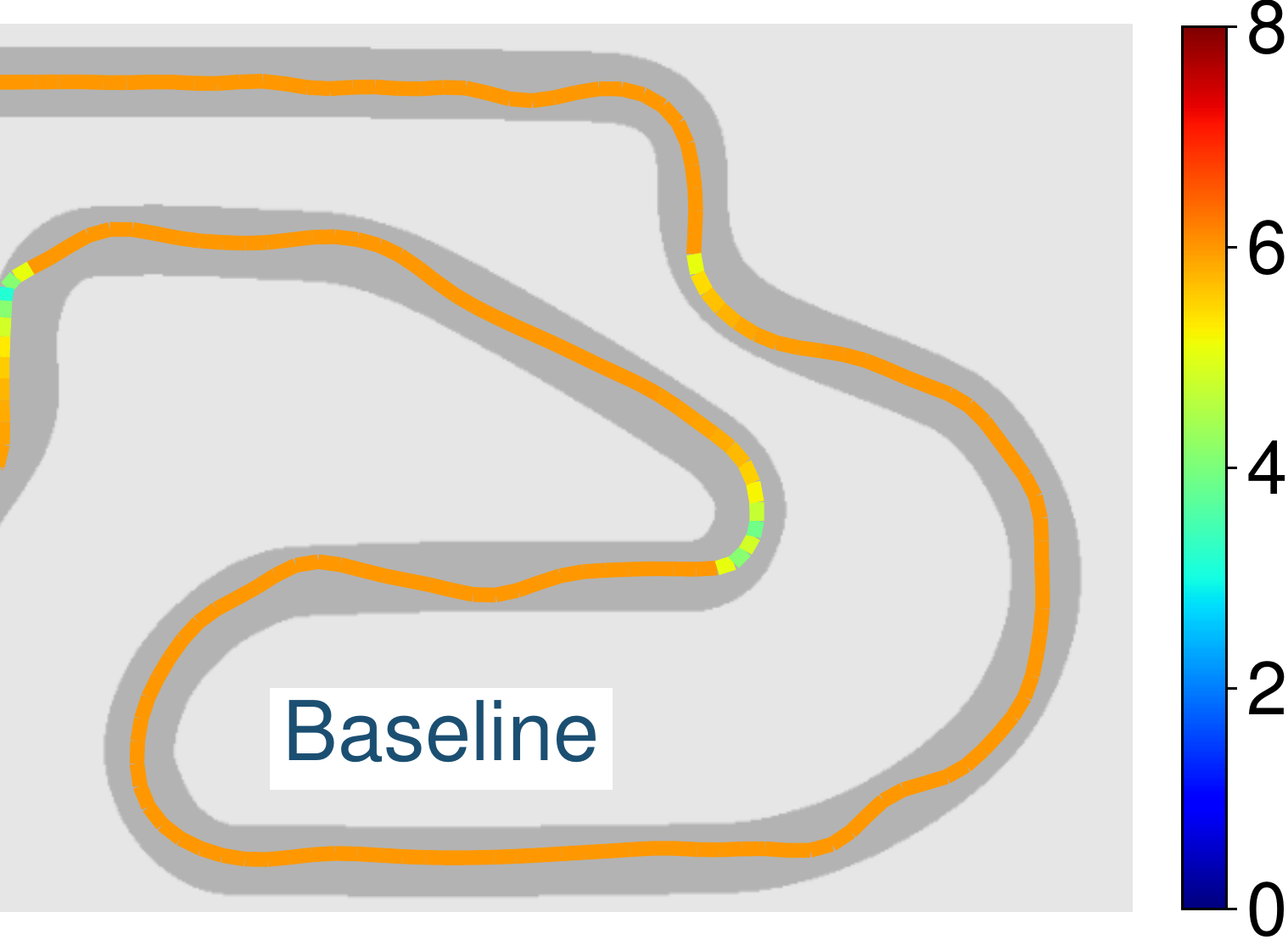}
    \end{minipage}
    \begin{minipage}{0.32\linewidth}
        \centering
        \includegraphics[width=\textwidth]{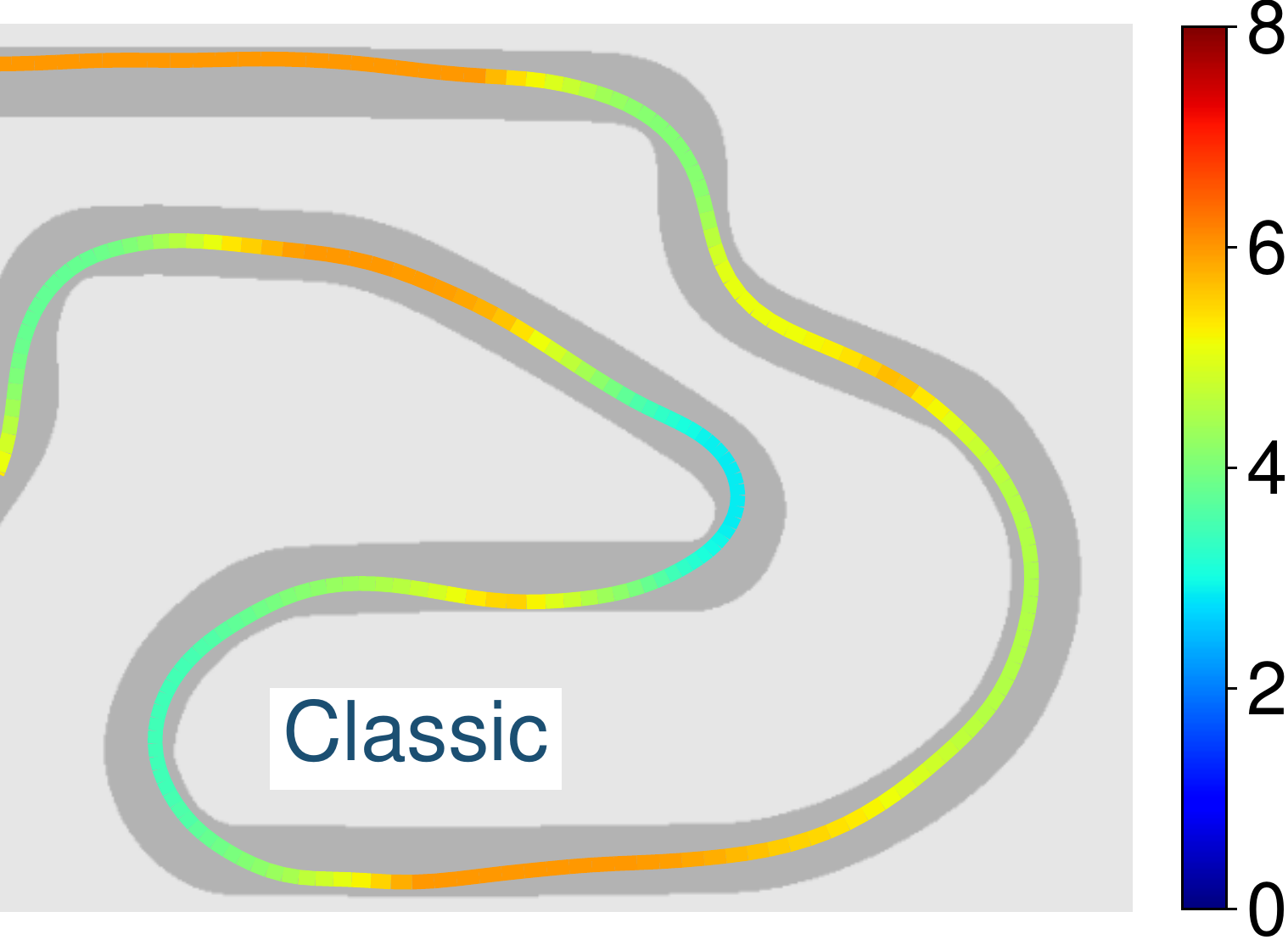}
    \end{minipage}
    \begin{minipage}{0.32\linewidth}
        \centering
        \includegraphics[width=\textwidth]{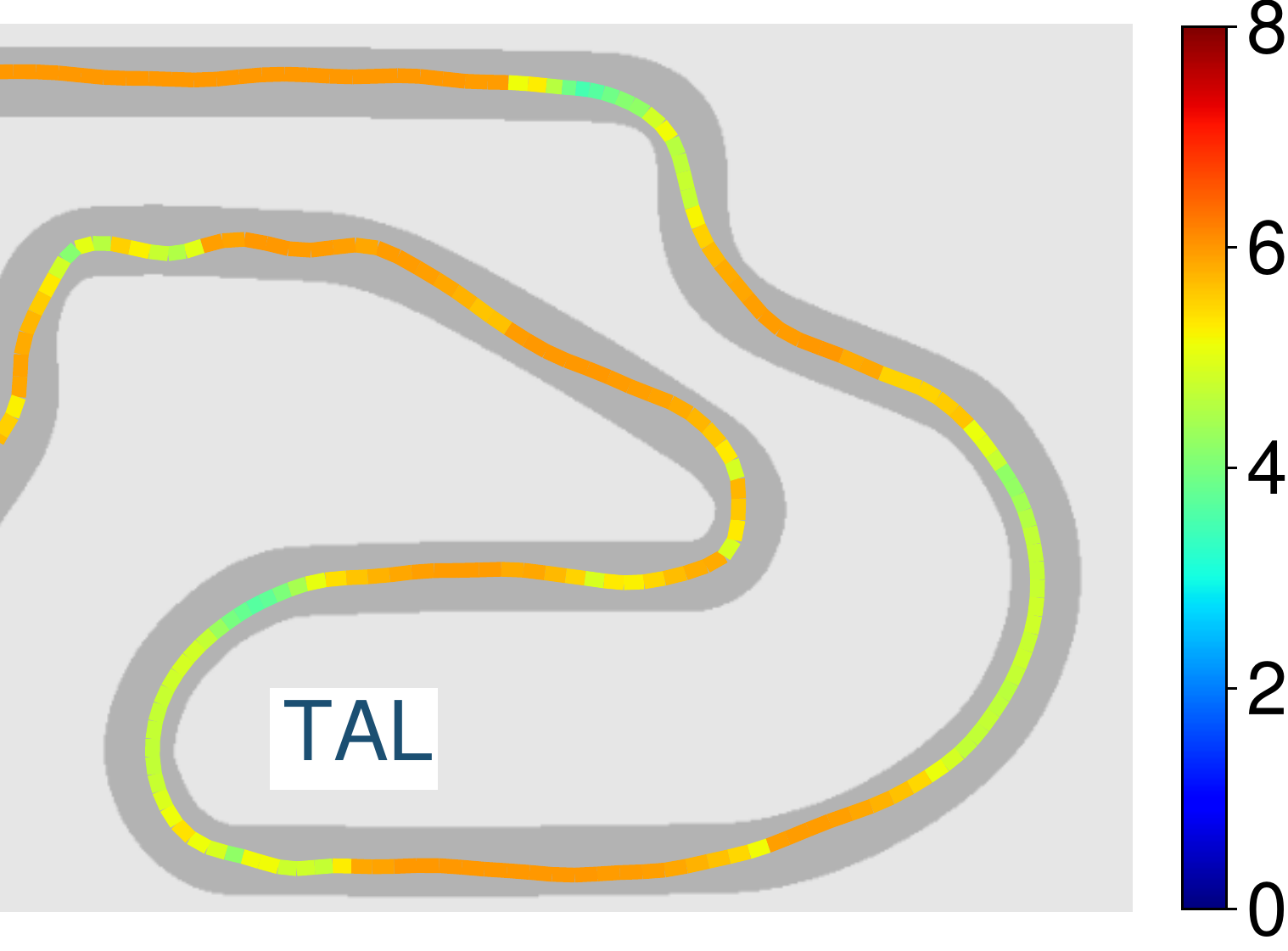}
    \end{minipage}
    \caption{Trajectories taken by the baseline (left), classic (middle) and TAL (right) planners on a portion of the ESP track.}
    \label{fig:trajectories_6ms}
\end{figure}

Fig. \ref{fig:trajectories_6ms} shows trajectories taken by the baseline, classic and TAL agents for a portion of the ESP track with a maximum speed of 6 m/s.
The baseline trajectory is mainly orange in both the straights and corners, indicating a near-constant speed of around 6 m/s for most of the trajectory.
In contrast, the classic trajectory has green, yellow and orange components indicating that the vehicle slows down in the corners and speeds up in the straights.
The TAL agent learns to select a similar speed profile to the classic planner of speeding up and slowing down.

\begin{figure}[h]
    \centering
    \includegraphics[width=0.9\linewidth]{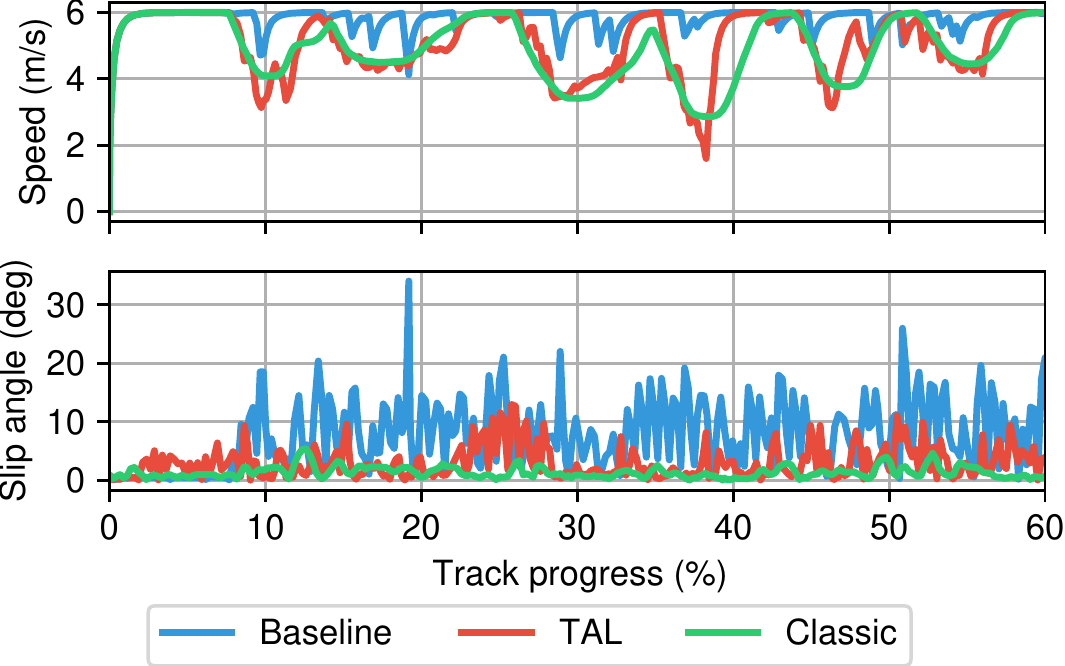}
    \caption{Speed and absolute slip angle for the baseline, TAL and classical planners on a portion of the ESP map.}
    \label{fig:speed_slip_profiles}
\end{figure}

Fig. \ref{fig:speed_slip_profiles} plots the speed and slip profiles of the baseline, TAL and classical planners for a portion of the ESP track. 
The speed graph confirms that the baseline planner selects high speeds near the maximum for most of the trajectory.
The classical planner smoothly slows down and speeds up, and the TAL agent approximately tracks the classical planner.

The bottom graph in Fig. \ref{fig:speed_slip_profiles} shows the corresponding absolute slip angles for the speed profiles.
The slip angle is the angle between the vehicle orientation and the direction of the velocity.
The classical planner has the smallest slip angle, followed by the TAL agent which reaches 10\degree.
The baseline agent has a significantly larger slip angle, regularly exceeding 15 \degree and reaching over 30\degree. 
This shows that the baseline agent relies on the vehicle drifting for much of the track, thus exploiting the simulation model.
This behaviour has been seen in other learning approaches \cite{zheng2022gradient, trumpp2023residual} and is responsible for causing the low completion rates.
Policies relying on high-slip angles in the simulator are not feasible for physical implementation since in reality tyre dynamics are non-linear and thus the policy learned in simulation differs from how the real-world vehicles perform.

\subsection{Performance Comparison - 8 m/s:}

We compare the TAL agent with a classical planner using the vehicle's maximum speed of 8 m/s.


\begin{figure}[h]
    \centering
    \begin{minipage}{0.32\linewidth}
        \centering
        \includegraphics[width=0.99\textwidth]{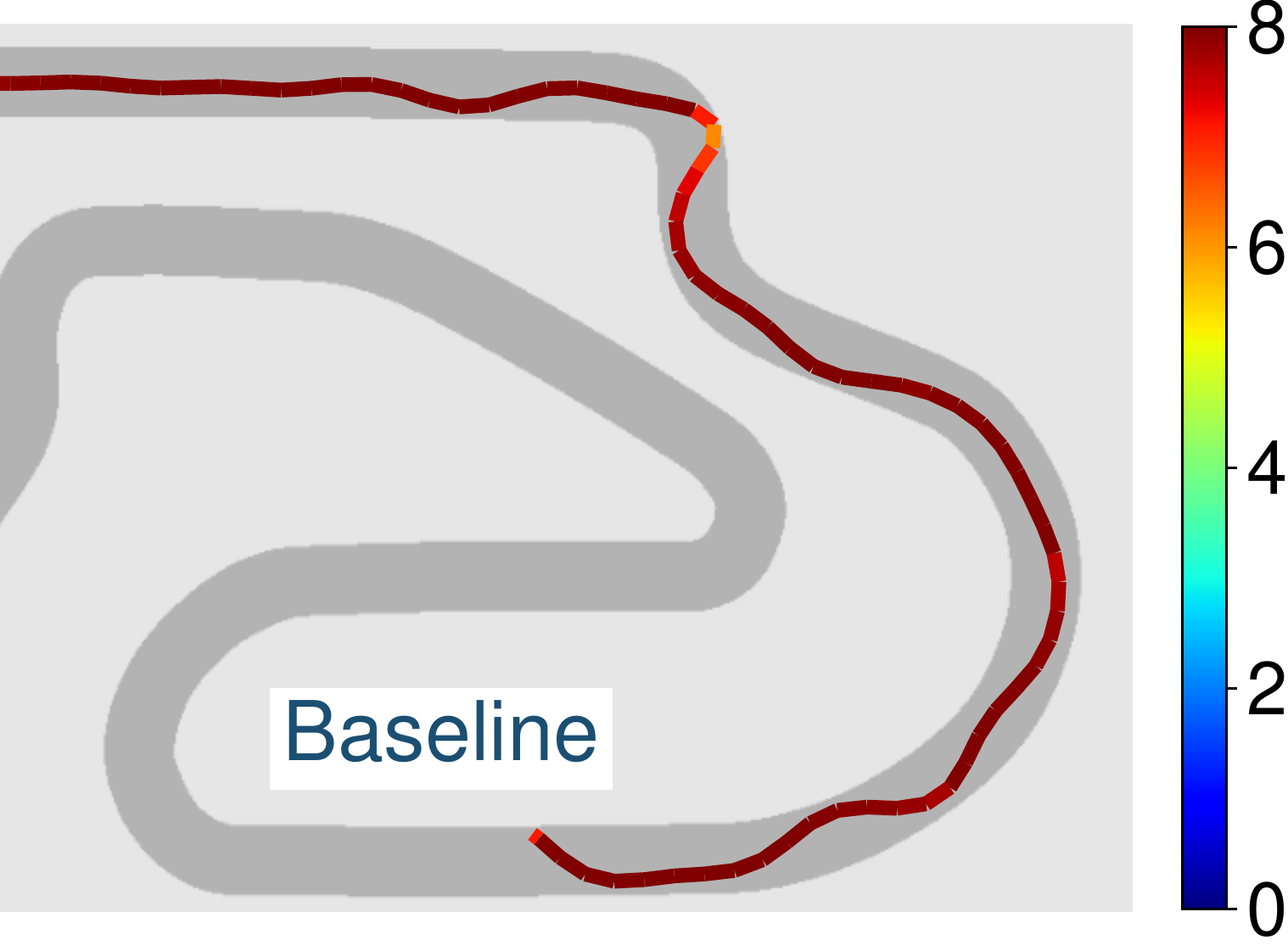}
    \end{minipage}
    \begin{minipage}{0.32\linewidth}
        \centering
        \includegraphics[width=0.99\textwidth]{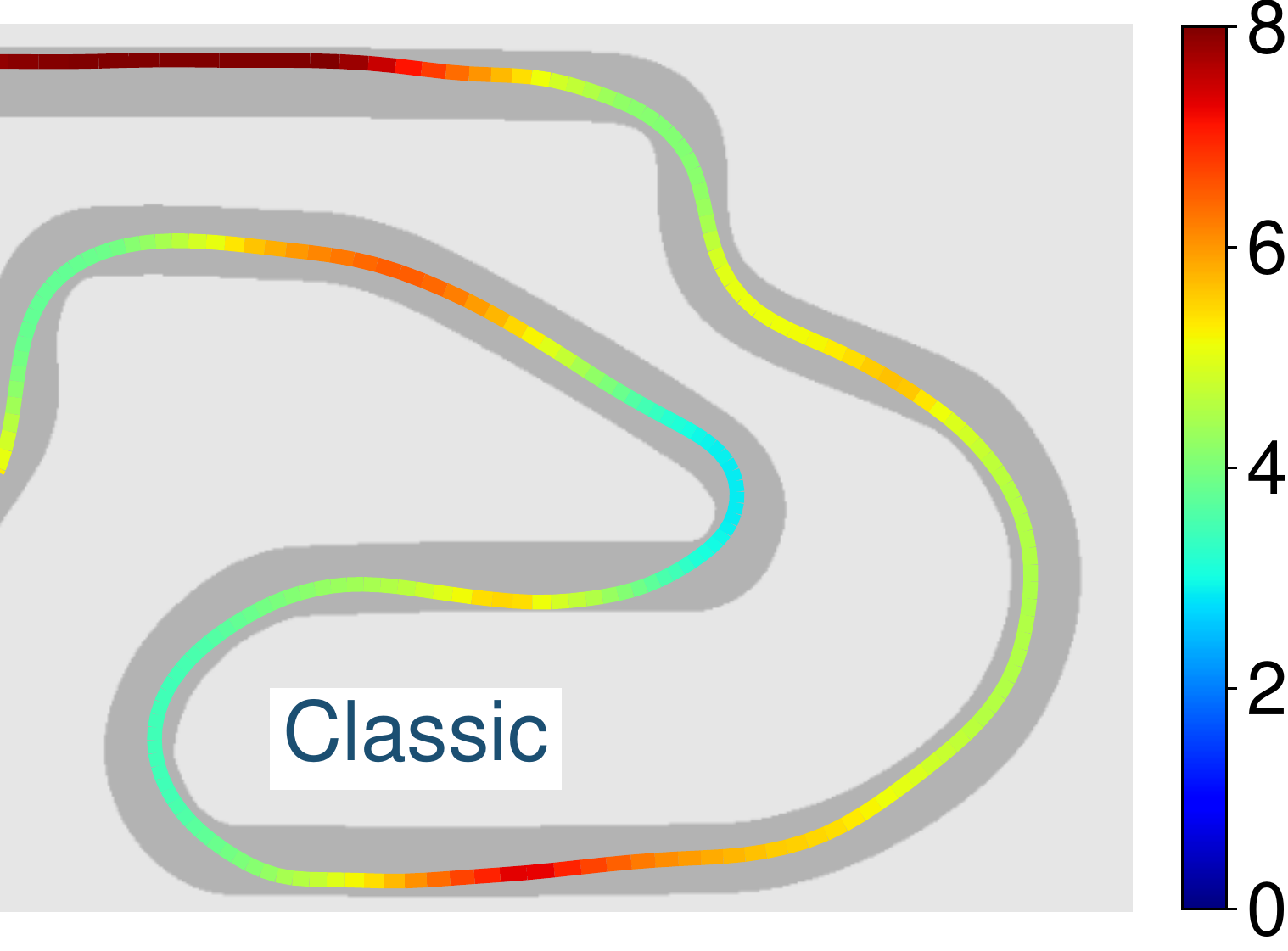}
    \end{minipage}
    \begin{minipage}{0.32\linewidth}
        \centering
        \includegraphics[width=0.99\textwidth]{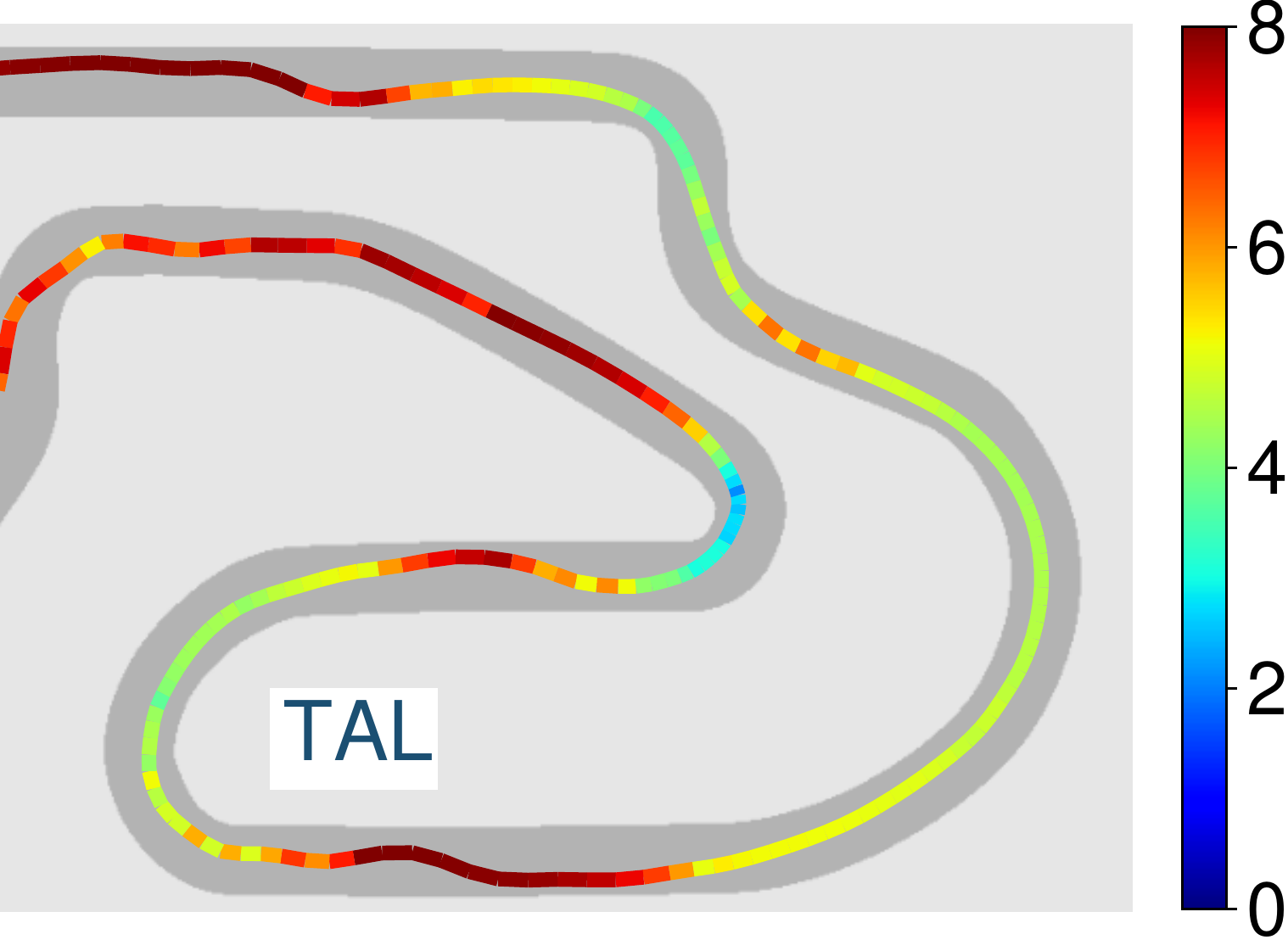}
    \end{minipage}
    \caption{Trajectories taken by the baseline (left), classical (middle) and TAL (right) agents on a portion of the ESP track.}
    \label{fig:baseline_tal_trajectories}
\end{figure}

Fig. \ref{fig:baseline_tal_trajectories} shows the trajectories selected by the baseline, classic and TAL planners.
The baseline agent selects near the maximum speed, resulting in the vehicle sliding and crashing early in the lap.
Following the racing line, the classic planner smoothly speeds up and slows down.
The TAL agent shows a similar pattern to the classic planner of speeding up in the straight sections and slowing down around the corners.

\begin{figure}[h]
    \centering
    \includegraphics[width=0.9\linewidth]{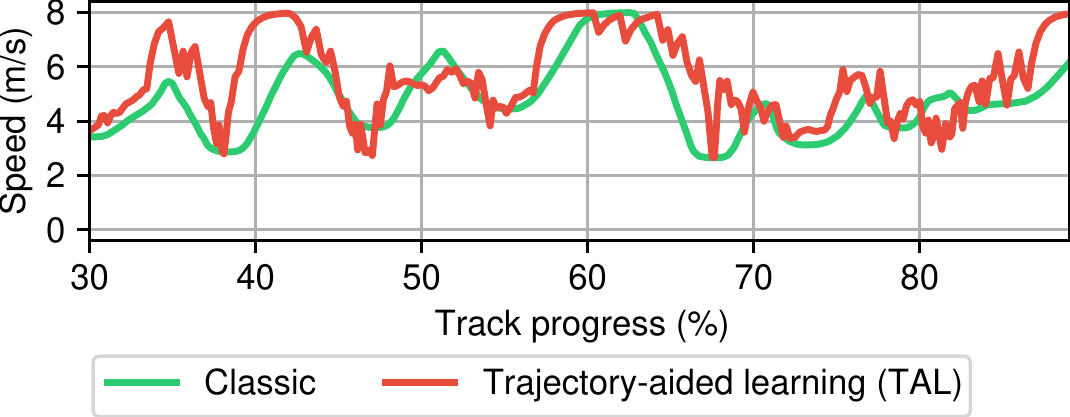}
    \caption{The speed profile of the classic planner and TAL agent using a maximum speed of 8 m/s on a portion of the ESP track.}
    \label{fig:speed_classical_tal_esp}
\end{figure}

Fig. \ref{fig:speed_classical_tal_esp} shows the speeds selected by the classical planner and TAL agent when both use the maximum speed of 8 m/s. 
The TAL agent roughly tracks the classical planner through the whole segment, occasionally deviating by selecting higher speeds or changing speed quickly.
The similar speed profiles show that the trajectory-aided learning formulation successfully trains the DRL agent to select a speed profile similar to the optimal trajectory.
A persisting limitation is that the DRL agent's actions are less smooth than the classical planner.

\begin{figure}[h]
    \centering
    \includegraphics[width=\linewidth]{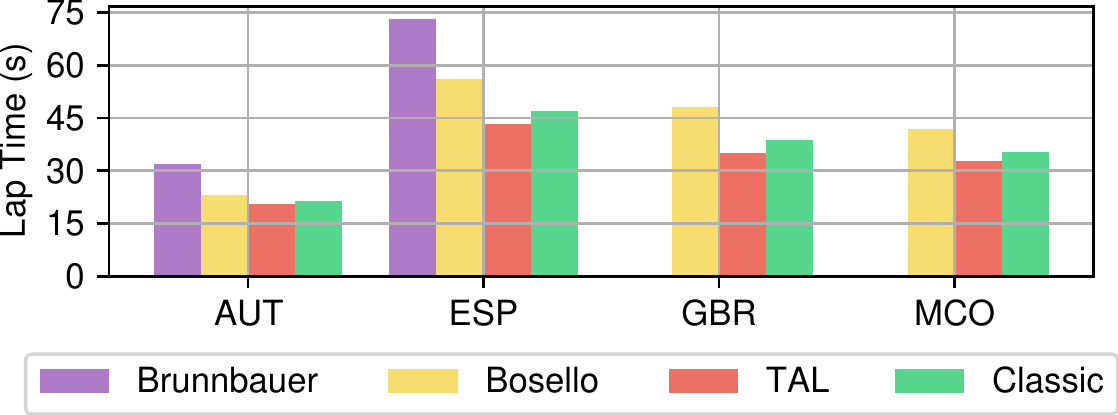}
    \caption{The lap times comparing the TAL agent to the results presented by Brunnbauer et al. \cite{Brunnbauer2022LatentRacing} and Bosello et al. \cite{Bosello2022TrainRaces} for the four test tracks.}
    \label{fig:literature_lap_times}
\end{figure}

Using the vehicle's maximum speed of 8 m/s, we compare the lap times from TAL agents to the classical planner and similar methods from the literature.
Fig. \ref{fig:literature_lap_times} shows the lap times achieved by the TAL agent compared to a classical planner and the results presented by Brunnbauer et al. \cite{Brunnbauer2022LatentRacing} and Bosello et al. \cite{Bosello2022TrainRaces}.
The classical and TAL planners use a maximum speed of 8 m/s, showing that a higher maximum speed allows them to complete laps faster than previous methods.
We, therefore, conclude that our approach trains agents to select better speed profiles, using higher maximum speeds and, therefore, better suited to autonomous racing than previous approaches.

\section{Conclusion}

This paper addressed the problem of training end-to-end DRL agents for high-speed racing.
We presented trajectory-aided learning, which rewards the agent according to the difference between the agent's actions and those selected by a classical planner following the optimal trajectory.
The evaluation showed that our proposed approach trains agents to race at high speeds with significantly higher completion rates than the baseline formulation.
Further investigation into the speed profile showed that this is due to the TAL agents selecting a better speed profile of slowing down in the corners and speeding up in the straights.
The improved speed profile causes the TAL agents to achieve a significantly higher completion rate on four test tracks.
Due to our approach using a higher maximum speed, the TAL agents achieve faster lap times than comparable methods in the literature.

The results in this paper demonstrate that incorporating classical components in the learning formulation improves the performance of DRL agents.
Using domain knowledge enables DRL agents to achieve good results in high-performance control.
Future work should study how these improvements to racing performance transfer to actual vehicles.
TAL agents are expected to transfer better to physical vehicles because they select appropriate speed profiles, thus having smaller slip angles.
Another extension of this work is using optimal trajectories in learning formulations for other applications such as drone control.


\typeout{}


\begin{thebibliography}{10}
\providecommand{\url}[1]{#1}
\csname url@samestyle\endcsname
\providecommand{\newblock}{\relax}
\providecommand{\bibinfo}[2]{#2}
\providecommand{\BIBentrySTDinterwordspacing}{\spaceskip=0pt\relax}
\providecommand{\BIBentryALTinterwordstretchfactor}{4}
\providecommand{\BIBentryALTinterwordspacing}{\spaceskip=\fontdimen2\font plus
\BIBentryALTinterwordstretchfactor\fontdimen3\font minus
    \fontdimen4\font\relax}
\providecommand{\BIBforeignlanguage}[2]{{%
\expandafter\ifx\csname l@#1\endcsname\relax
\typeout{** WARNING: IEEEtran.bst: No hyphenation pattern has been}%
\typeout{** loaded for the language `#1'. Using the pattern for}%
\typeout{** the default language instead.}%
\else
\language=\csname l@#1\endcsname
\fi
#2}}
\providecommand{\BIBdecl}{\relax}
\BIBdecl

\bibitem{Betz2022AutonomousRacing}
J.~Betz, H.~Zheng, A.~Liniger, U.~Rosolia, P.~Karle, M.~Behl, V.~Krovi, and
    R.~Mangharam, ``Autonomous vehicles on the edge: A survey on autonomous
    vehicle racing,'' \emph{IEEE Open Journal of Intelligent Transportation
    Systems}, 2022.

\bibitem{wang2021data}
R.~Wang, ``Data-driven system identification and optimal control framework for
    grand-prix style autonomous racing,'' Ph.D. dissertation, Clemson University,
    2021.

\bibitem{Heilmeier2020MinimumCar}
A.~Heilmeier, A.~Wischnewski, L.~Hermansdorfer, J.~Betz, M.~Lienkamp, and
    B.~Lohmann, ``{Minimum curvature trajectory planning and control for an
    autonomous race car},'' \emph{Vehicle System Dynamics}, vol.~58, no.~10, pp.
    1497--1527, 10 2020.

\bibitem{WalshCDDTLocalization}
C.~H. Walsh and S.~Karaman, ``Cddt: Fast approximate 2d ray casting for
    accelerated localization,'' in \emph{2018 IEEE International Conference on
    Robotics and Automation (ICRA)}.\hskip 1em plus 0.5em minus 0.4em\relax IEEE,
    2018, pp. 3677--3684.

\bibitem{zhang2016learning}
T.~Zhang, G.~Kahn, S.~Levine, and P.~Abbeel, ``Learning deep control policies
    for autonomous aerial vehicles with mpc-guided policy search,'' in \emph{2016
    IEEE international conference on robotics and automation (ICRA)}.\hskip 1em
    plus 0.5em minus 0.4em\relax IEEE, 2016, pp. 528--535.

\bibitem{sutton2018reinforcement}
R.~S. Sutton and A.~G. Barto, \emph{Reinforcement learning: An
    introduction}.\hskip 1em plus 0.5em minus 0.4em\relax MIT press, 2018.

\bibitem{hamilton2022zero}
N.~Hamilton, P.~Musau, D.~M. Lopez, and T.~T. Johnson, ``Zero-shot policy
    transfer in autonomous racing: Reinforcement learning vs imitation
    learning,'' in \emph{2022 IEEE International Conference on Assured Autonomy
    (ICAA)}.\hskip 1em plus 0.5em minus 0.4em\relax IEEE, 2022, pp. 11--20.

\bibitem{ivanov2020case}
R.~Ivanov, T.~J. Carpenter, J.~Weimer, R.~Alur, G.~J. Pappas, and I.~Lee,
    ``Case study: verifying the safety of an autonomous racing car with a neural
    network controller,'' in \emph{Proceedings of the 23rd International
    Conference on Hybrid Systems: Computation and Control}, 2020, pp. 1--7.

\bibitem{Bosello2022TrainRaces}
M.~Bosello, R.~Tse, and G.~Pau, ``Train in austria, race in montecarlo:
    Generalized rl for cross-track f1 tenth lidar-based races,'' in \emph{2022
    IEEE 19th Annual Consumer Communications \& Networking Conference
    (CCNC)}.\hskip 1em plus 0.5em minus 0.4em\relax IEEE, 2022, pp. 290--298.

\bibitem{wischnewski2022indy}
A.~Wischnewski, M.~Geisslinger, J.~Betz, T.~Betz, F.~Fent, A.~Heilmeier,
    L.~Hermansdorfer, T.~Herrmann, S.~Huch, P.~Karle \emph{et~al.}, ``Indy
    autonomous challenge-autonomous race cars at the handling limits,'' in
    \emph{12th International Munich Chassis Symposium 2021}.\hskip 1em plus 0.5em
    minus 0.4em\relax Springer, 2022, pp. 163--182.

\bibitem{o2020tunercar}
M.~O’Kelly, H.~Zheng, A.~Jain, J.~Auckley, K.~Luong, and R.~Mangharam,
    ``Tunercar: A superoptimization toolchain for autonomous racing,'' in
    \emph{2020 IEEE International Conference on Robotics and Automation
    (ICRA)}.\hskip 1em plus 0.5em minus 0.4em\relax IEEE, 2020, pp. 5356--5362.

\bibitem{tuatulea2020design}
A.~T{\u{a}}tulea-Codrean, T.~Mariani, and S.~Engell, ``Design and simulation of
    a machine-learning and model predictive control approach to autonomous race
    driving for the f1/10 platform,'' \emph{IFAC-PapersOnLine}, vol.~53, no.~2,
    pp. 6031--6036, 2020.

\bibitem{coulter1992implementation}
R.~C. Coulter, ``Implementation of the pure pursuit path tracking algorithm,''
    Carnegie-Mellon UNIV Pittsburgh PA Robotics INST, Tech. Rep., 1992.

\bibitem{becker2022model}
J.~Becker, N.~Imholz, L.~Schwarzenbach, E.~Ghignone, N.~Baumann, and M.~Magno,
    ``Model-and acceleration-based pursuit controller for high-performance
    autonomous racing,'' \emph{arXiv preprint arXiv:2209.04346}, 2022.

\bibitem{chisari2021learning}
E.~Chisari, A.~Liniger, A.~Rupenyan, L.~Van~Gool, and J.~Lygeros, ``Learning
    from simulation, racing in reality,'' in \emph{2021 IEEE International
    Conference on Robotics and Automation (ICRA)}.\hskip 1em plus 0.5em minus
    0.4em\relax IEEE, 2021, pp. 8046--8052.

\bibitem{Cai2020HighSpeedLearning}
P.~Cai, X.~Mei, L.~Tai, Y.~Sun, and M.~Liu, ``{High-Speed Autonomous Drifting
    with Deep Reinforcement Learning},'' \emph{IEEE Robotics and Automation
    Letters}, vol.~5, no.~2, pp. 1247--1254, 4 2020.

\bibitem{ghignone2022tc}
E.~Ghignone, N.~Baumann, M.~Boss, and M.~Magno, ``Tc-driver: Trajectory
    conditioned driving for robust autonomous racing--a reinforcement learning
    approach,'' \emph{arXiv preprint arXiv:2205.09370}, 2022.

\bibitem{dwivedi2022continuous}
T.~Dwivedi, T.~Betz, F.~Sauerbeck, P.~Manivannan, and M.~Lienkamp, ``Continuous
    control of autonomous vehicles using plan-assisted deep reinforcement
    learning,'' in \emph{2022 22nd International Conference on Control,
    Automation and Systems (ICCAS)}.\hskip 1em plus 0.5em minus 0.4em\relax IEEE,
    2022, pp. 244--250.

\bibitem{jaritz2018end}
M.~Jaritz, R.~De~Charette, M.~Toromanoff, E.~Perot, and F.~Nashashibi,
    ``End-to-end race driving with deep reinforcement learning,'' in \emph{2018
    IEEE International Conference on Robotics and Automation (ICRA)}.\hskip 1em
    plus 0.5em minus 0.4em\relax IEEE, 2018, pp. 2070--2075.

\bibitem{evans2022accelerating}
B.~Evans, J.~Betz, H.~Zheng, H.~A. Engelbrecht, R.~Mangharam, and H.~W.
    Jordaan, ``Accelerating online reinforcement learning via supervisory safety
    systems,'' \emph{arXiv preprint arXiv:2209.11082}, 2022.

\bibitem{sun2022benchmark}
X.~Sun, M.~Zhou, Z.~Zhuang, S.~Yang, J.~Betz, and R.~Mangharam, ``A benchmark
    comparison of imitation learning-based control policies for autonomous
    racing,'' \emph{arXiv preprint arXiv:2209.15073}, 2022.

\bibitem{musau2022using}
P.~Musau, N.~Hamilton, D.~M. Lopez, P.~Robinette, and T.~T. Johnson, ``On using
    real-time reachability for the safety assurance of machine learning
    controllers,'' in \emph{2022 IEEE International Conference on Assured
    Autonomy (ICAA)}.\hskip 1em plus 0.5em minus 0.4em\relax IEEE, 2022, pp.
    1--10.

\bibitem{Brunnbauer2022LatentRacing}
A.~Brunnbauer, L.~Berducci, A.~Brandstatter, M.~Lechner, R.~Hasani, D.~Rus, and
    R.~Grosu, ``{Latent Imagination Facilitates Zero-Shot Transfer in Autonomous
    Racing},'' \emph{2022 International Conference on Robotics and Automation
    (ICRA)}, pp. 7513--7520, 5 2022.

\bibitem{zhang2022residual}
R.~Zhang, J.~Hou, G.~Chen, Z.~Li, J.~Chen, and A.~Knoll, ``Residual policy
    learning facilitates efficient model-free autonomous racing,'' \emph{IEEE
    Robotics and Automation Letters}, vol.~7, no.~4, pp. 11\,625--11\,632, 2022.

\bibitem{Lillicrap2016ContinuousLearning}
T.~P. Lillicrap, J.~J. Hunt, A.~Pritzel, N.~Heess, T.~Erez, Y.~Tassa,
    D.~Silver, and D.~Wierstra, ``{Continuous control with deep reinforcement
    learning},'' \emph{4th International Conference on Learning Representations,
    ICLR 2016 - Conference Track Proceedings}, 2016.

\bibitem{fujimoto2018addressing}
S.~Fujimoto, H.~Hoof, and D.~Meger, ``Addressing function approximation error
    in actor-critic methods,'' in \emph{International conference on machine
    learning}.\hskip 1em plus 0.5em minus 0.4em\relax PMLR, 2018, pp. 1587--1596.

\bibitem{o2020f1tenth}
M.~O'Kelly, H.~Zheng, D.~Karthik, and R.~Mangharam, ``F1tenth: An open-source
    evaluation environment for continuous control and reinforcement learning,''
    \emph{Proceedings of Machine Learning Research}, vol. 123, 2020.

\bibitem{Althoff2017}
M.~Althoff, M.~Koschi, and S.~Manzinger, ``{CommonRoad}: Composable benchmarks
    for motion planning on roads,'' in \emph{2017 {IEEE} Intelligent Vehicles
    Symposium ({IV})}.\hskip 1em plus 0.5em minus 0.4em\relax {IEEE}, Jun. 2017.

\bibitem{zheng2022gradient}
H.~Zheng, J.~Betz, and R.~Mangharam, ``Gradient-free multi-domain optimization
    for autonomous systems,'' \emph{arXiv preprint arXiv:2202.13525}, 2022.

\bibitem{trumpp2023residual}
R.~Trumpp, D.~Hoornaert, and M.~Caccamo, ``Residual policy learning for vehicle
    control of autonomous racing cars,'' \emph{arXiv preprint arXiv:2302.07035},
    2023.

\end{thebibliography}

\end{document}